\documentclass[12pt]{iopart}

\usepackage{iopams}  

\usepackage{cite}
\usepackage{graphicx}
\usepackage{bm}
\usepackage{color}

\newcommand{\argmin}{\mathop{\rm arg~min}\limits}

\begin{document}

\title{Koopman operator-based discussion on partial observation in stochastic systems}

\author{Jun Ohkubo}

\address{Graduate School of Science and Engineering, Saitama University, \\
255 Shimo-Okubo, Sakura-ku, Saitama 338-8570, Japan}
\ead{johkubo@mail.saitama-u.ac.jp}

\vspace{10pt}

\begin{abstract}
It is sometimes difficult to achieve a complete observation for a full set of observables, and partial observations are necessary. For deterministic systems, the Mori-Zwanzig formalism provides a theoretical framework for handling partial observations. Recently, data-driven algorithms based on the Koopman operator theory have made significant progress, and there is a discussion to connect the Mori-Zwanzig formalism with the Koopman operator theory. In this work, we discuss the effects of partial observation in stochastic systems using the Koopman operator theory. The discussion clarifies the importance of distinguishing the state space and the function space in stochastic systems. Even in stochastic systems, the delay-embedding technique is beneficial for partial observation, and several numerical experiments show a power-law behavior of error with respect to the amplitude of the additive noise. We also discuss the relation between the exponent of the power-law behavior and the effects of partial observation.
\end{abstract}

%
%
%
%
%

\section{Introduction}

In a classical dynamical system, the behavior is deterministic if one can observe complete information about the system. However, it would be difficult to perform a perfect observation, and a partial observation causes stochasticity. A simple example is the movement of a ball; if only the coordinate is measurable but not its velocity, it is impossible to predict in which direction and how much it will move at the next time. The effect of partial observation has also been discussed in statistical physics; constructing coarse-grained models from high-dimensional microscopic ones leads to time-evolution equations of a smaller set of relevant variables of interest. The unobserved variables could play a role as noise. In the recent development of data-driven approaches, the effects of partial observation are crucial because it would be difficult to observe all variables. Hence, it would be desirable to obtain even a portion of the overall information from partial observation.

One of the famous ways for handling partial observations is the Mori-Zwanzig formalism \cite{Zwanzig1961,Mori1965,Zwanzig1973}. Mori and Zwanzig developed projection methods to express the effect of unobserved variables in terms of observed ones. The time-evolution equation for the observed variables is called the generalized Langevin equation. The crucial points in the generalized Langevin equation are as follows:

\begin{itemize}
\item Although the original dynamics is deterministic and Markovian, the generalized Langevin equation has a memory term that depends on the history.
\item There is a term that could be interpreted as noise; the noise term represents the orthogonal dynamics and depends on the unknown initial conditions of the unobserved variables.
\end{itemize}
There are many works on the Mori-Zwanzig formalism. For example, some employed the Mori-Zwanzig formalism to improve prediction accuracy \cite{Chorin2000,Chorin2002,Chorin2006,Gouasmi2017,teVrugt2024,Netz2024}. There are many related works; see, for example, the references in \cite{Netz2024}. Some works focused on data-driven approaches for the Mori-Zwanzig formalism \cite{Chorin2015,Meyer2019,Maeyama2020,Gonzalez2021,Tian2021}, and some discussed its application in neural networks \cite{Venturi2023,Gupta2025}.

Recently, a data-driven approach based on the Koopman operator has been connected to the discussion of the Mori-Zwanzig formalism \cite{LinKK2021,LinYT2021,Lin2023}. The Koopman operator \cite{Koopman1931} enables us to deal with nonlinear dynamical systems in terms of linear algebra. The well-known methods of incorporating data into the Koopman operator include the dynamic mode decomposition (DMD)\cite{Rowley2009,Schmid2010}, the extended dynamic mode decomposition (EDMD)\cite{Williams2015}, and the Hankel DMD\cite{Arbabi2017,Brunton2017}. For details of the Koopman approach, see the book \cite{Mauroy2020} and reviews \cite{Budisic2012,Mezic2013,Rowley2017,Mezic2021,Brunton2022}. In \cite{LinYT2021}, a discussion based on the Koopman approach naturally leads to the generalized Langevin equation and algorithms for estimating the key components from time series datasets. There are also some works on this topic; \cite{LinKK2021} focused on the Wiener projection, and \cite{Lin2023} employed regression-based projections.

Here, we note that previous discussions on the Koopman-based approach to partial observation have been basically restricted only to deterministic dynamical systems. Of course, the original Mori-Zwanzig formalism targeted deterministic systems, and it would be natural to consider the Koopman-based approach to deterministic cases. However, the Koopman approach is not restricted to deterministic cases and is available to stochastic systems; for example, see \cite{Williams2015,Crnjaric-Zic2020,Wanner2022}. In some cases, one would consider time-evolution equations with noise as the starting point. Here, note that the noise term in the generalized Langevin equation stems from the effects of unobserved variables. Then, how does the inherent probabilistic nature of probability theory affect this discussion? For example, stochastic differential equations with additive Wiener noise are widely used in statistical physics, and it remains unclear how to consider partial observations in stochastic differential equations.

In the present paper, we discuss the effects of partial observation in stochastic systems. Here, we focus on the stochastic differential equation with an additive Wiener noise. To discuss partial observation, the Koopman operator approach is employed. In the Koopman operator approach, it is crucial to distinguish the state space and the function space; we will clarify that the generalized Langevin equation in the state space is derived only in deterministic cases. The discussion on stochastic systems clarifies the effect of delay-embedding, which improves estimation accuracy over the use of higher-order basis functions. In addition, numerical experiments for the noisy van der Pol system and the noise Lorenz system show a power-law dependency of the accuracy on the noise amplitude in the partial observation settings.

This article is organized as follows. In Sec.~2, we review the Koopman operator approach. Section~3 briefly summarizes the previous discussions on the Mori-Zwanzig formalism with the Koopman operator approach, where we will emphasize the role of the deterministic feature in the explanation. Then, Sec.~4 yields discussions on the partial observations in stochastic systems and the effects of delay-embedding. We also present some results from numerical experiments. In the numerical experiments, the noisy van der Pol system and the noisy Lorenz system are employed as toy examples.

\section{Preliminaries on Koopman operator approach}

In the present paper, it is crucial to distinguish the state vectors and observable functions that yield the values of the state vectors. We review the Koopman operator approach with an emphasis on this point. For further details, refer to \cite{Williams2015}.

\subsection{Distinction between state space and function space}

Consider a state space $\mathcal{M} \subseteq \mathbb{R}^{D}$ and an adequate functional space $\mathcal{F}$; an element in $\mathcal{F}$ is called an observable function, and $\mathcal{F} \ni \phi: \mathcal{M} \to \mathbb{R}$. To emphasize the distinctions between the state space and the functional space, we add underlines to denote state vectors on $\mathcal{M}$, e.g., $\underline{\bm{x}}$. The $d$-th element of $\underline{\bm{x}}$ is denoted as $\underline{x_d} \in \mathbb{R}$. By contrast, an observable function, which yields the value of the $d$-th element $\underline{x_d}$, is denoted as $x_d(\underline{\bm{x}}) = \underline{x_d}$. Note that it is common in many papers on the Koopman operator approach to use abbreviations, e.g., $x_d \in \mathcal{F}$, for notational simplicity. While the abbreviations are convenient, we again emphasize the importance of distinguishing between elements in the state space and those in the function space. Hence, we employ the notation with underlines for the state space.

\subsection{Deterministic systems}

Let $\underline{\bm{x}(t)}$ be the state vector of the system at time $t$. The deterministic time-evolution equation for $\underline{\bm{x}(t)}$ is given by
\begin{eqnarray}
\frac{\rmd}{\rmd t} \underline{\bm{x}(t)} = \bm{a}(\underline{\bm{x}(t)}),
\label{eq_ode}
\end{eqnarray}
where $\bm{a}(\underline{\bm{x}})$ is a vector of functions giving the time-evolution for each coordinate $\underline{x_d(t)}$. While it could be possible to include the time variable $t$ in the functions, we assume that $\bm{a}(\underline{\bm{x}})$ is time-independent for simplicity.

It is beneficial to see the coupled ordinary differential equations in \eref{eq_ode} from the viewpoint of a dynamical system with discrete-time steps. Setting a time interval for observation as $\Delta t_{\mathrm{obs}}$, the state vector evolves in time as follows:
\begin{eqnarray}
\underline{\bm{x}(t+\Delta t_{\mathrm{obs}})} = \bm{F}_{\Delta t_{\mathrm{obs}}} (\underline{\bm{x}(t)}),
\label{eq_state_space_evolution}
\end{eqnarray}
where $\bm{F}_{\Delta t_{\mathrm{obs}}}: \mathcal{M} \to \mathcal{M}$ yields the state vector after the time-evolution with $\Delta t_{\mathrm{obs}}$.

In the Koopman operator approach, we consider the time-evolution of observable functions instead of that of the state vector. Consider the observable function $x_d: \mathcal{M} \to \mathbb{R}$, which yields the $d$-th element of the state vector. Then, we consider a function that gives the $d$-th element of the state vector after the time-evolution with $\Delta t_{\mathrm{obs}}$. As we will see later, it is possible to obtain such a function by employing a map $\mathcal{K}_{\Delta t_{\mathrm{obs}}}: \mathcal{F} \to \mathcal{F}$. The map $\mathcal{K}_{\Delta t_{\mathrm{obs}}}$ is called the Koopman operator, and its action is defined as
\begin{eqnarray}
\mathcal{K}_{\Delta t_{\mathrm{obs}}} x_d = x_d \circ \bm{F}_{\Delta t_{\mathrm{obs}}},
\end{eqnarray}
which leads to
\begin{eqnarray}
x_d(\underline{\bm{x}(t+\Delta t_{\mathrm{obs}})}) = \mathcal{K}_{\Delta t_{\mathrm{obs}}} x_d(\underline{\bm{x}(t)}).
\end{eqnarray}
Note that the Koopman operator is linear even if the original system is nonlinear.

\begin{figure}[t]
\begin{center}
\includegraphics[width=100mm]{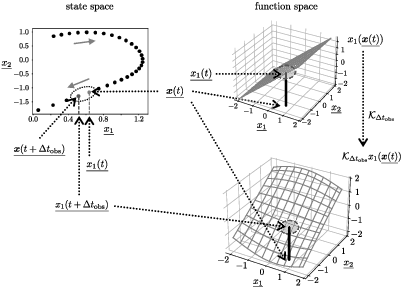}
\end{center}
\caption{The role of the Koopman operator $\mathcal{K}_{\Delta t_{\mathrm{obs}}}$. We here consider an example for $D=2$. The change of $\underline{x_1(t)}$ to $\underline{x_1(t+\Delta t_{\mathrm{obs}})}$ on the state space corresponds to $x_1(\underline{\bm{x}(t)})$ to $\mathcal{K}_{\Delta t_{\mathrm{obs}}}x_1(\underline{\bm{x}(t)})$ on the function space. Note that the Koopman operator varies the flat plane $x_1$ to the curved surface, which corresponds to the map on the function space $\mathcal{K}_{\Delta t_{\mathrm{obs}}}: \mathcal{F} \to \mathcal{F}$.}
\label{fig1}
\end{figure}

Figure~\ref{fig1} shows the correspondence between the state space $\mathcal{M}$ and the function space $\mathcal{F}$. Although the observable function $x_1$ is the simple flat plane, the action of the Koopman operator yields the curved surface $\mathcal{K}_{\Delta t_{\mathrm{obs}}}x_1$. The function $\mathcal{K}_{\Delta t_{\mathrm{obs}}}x_1: \mathcal{M} \to \mathbb{R}$ gives the coordinate value after the time-evolution, i.e., $\underline{x_1(t+\Delta t_{\mathrm{obs}})}$.

Note that while it is possible to consider various types of observable functions, we focus only on the simple observables $\{x_d\}$ in the present work. A collection of the observable functions for the coordinates is called a full-state observable $\bm{g}: \mathcal{M} \to \mathbb{R}^D$, which is a \textit{vector-valued} observable function and
\begin{eqnarray}
\bm{g} = [x_1 \quad x_2 \quad \dots \quad x_D]^{\top}.
\end{eqnarray}
Again, we note that it is crucial to distinguish $\bm{g}$ and $\underline{\bm{x}} = [\underline{x_1} \,\, \underline{x_2}\,\,\dots\,\,\underline{x_D}]^{\top} \in \mathbb{R}^D$.

\subsection{Extended dynamic mode decomposition}

As discussed in \cite{Williams2015}, we can obtain an approximate representation of the Koopman operator $\mathcal{K}_{\Delta t_{\mathrm{obs}}}$ by a data-driven approach. That is, the introduction of a set of basis functions, called a dictionary, leads to a Koopman matrix $K_{\Delta t_{\mathrm{obs}}}$ for the time interval $\Delta t_{\mathrm{obs}}$. The data-driven method is called the EDMD algorithm.

The EDMD algorithm requires a data set of snapshot pairs. The snapshot pairs are denoted as $\{ (\underline{\bm{x}_n}, \underline{\bm{y}_n}) \}_{n=1}^{N_\mathrm{data}}$, where 
\begin{eqnarray}
\underline{\bm{x}_n} =
[
\underline{x_{n1}(t)} \quad \underline{x_{n2}(t)} \quad \cdots \quad \underline{x_{nD}(t)}
]^{\top}
\end{eqnarray}
and 
\begin{eqnarray}
\underline{\bm{y}_n} &= \bm{F}_{\Delta t_{\mathrm{obs}}}(\underline{\bm{x}_n}) \nonumber \\
&= [\underline{x_{n1}(t+\Delta t_{\mathrm{obs}})} \quad \underline{x_{n2}(t+\Delta t_{\mathrm{obs}})} \quad \cdots \quad \underline{x_{nD}(t+\Delta t_{\mathrm{obs}})}]^{\top}.
\end{eqnarray}
$N_\mathrm{data}$ represents the number of snapshot pairs. The dictionary, $\bm{\psi}$, consists of $N_{\mathrm{dic}}$ basis functions, i.e.,
\begin{eqnarray}
\bm{\psi}(\underline{\bm{x}})
=
[
\psi_1(\underline{\bm{x}}) \quad \psi_2(\underline{\bm{x}}) \quad \cdots \quad \psi_{N_\mathrm{dic}}(\underline{\bm{x}})
]^{\top},
\end{eqnarray}
where $\psi_i: \mathcal{M} \to \mathbb{R}$ for $i = 1,\dots,N_\mathrm{dic}$. For example, a dictionary with monomial functions for $D=2$ is given as
\begin{eqnarray}
\fl
\bm{\psi}(\underline{\bm{x}})
=
[
\underline{1} \quad x_1(\underline{\bm{x}}) \quad x_2(\underline{\bm{x}})
\quad x_1^2(\underline{\bm{x}}) \quad x_1 x_2(\underline{\bm{x}}) \quad x_2^2(\underline{\bm{x}})
\quad x_1^3(\underline{\bm{x}}) \quad \dots \quad x_2^5(\underline{\bm{x}})
]^{\top},
\label{eq_monomial_dictionary_example}
\end{eqnarray}
where the maximum degree of the monomial functions is $5$, and $\underline{1}$ means a scalar value $1$. The dictionary spans a subspace $\mathcal{F}_{\bm{\psi}} \subseteq \mathcal{F}$.

Note that the dictionary with the monomial functions contains the full-state observable $\bm{g}$; see the second and third elements on the right-hand side of \eref{eq_monomial_dictionary_example}. However, the action of the Koopman operator yields the function $\mathcal{K}_{\Delta t_{\mathrm{obs}}} x_d$ that may not be in the spanned subspace $\mathcal{F}_{\bm{\psi}}$. In this case, we have the following expansion:
\begin{eqnarray}
\mathcal{K}_{\Delta t_{\mathrm{obs}}} x_d(\underline{\bm{x}}) \simeq \sum_{i=1}^{N_\mathrm{dic}} c^{\mathcal{K}_{\Delta t_{\mathrm{obs}}} x_d}_i \psi_i(\underline{\bm{x}}),
\label{eq_Kxd_basis_expansion}
\end{eqnarray}
where $\{c^{\mathcal{K}_{\Delta t_{\mathrm{obs}}}x_d}_i\}_{i=1}^{N_\mathrm{dic}}$ are the expansion coefficients. Similarly, we can consider the action on an arbitrary function $\phi \in \mathcal{F}$, $\mathcal{K}_{\Delta t_{\mathrm{obs}}} \phi$, and its coefficients $\{c^{\mathcal{K}_{\Delta t_{\mathrm{obs}}} \phi}_i\}_{i=1}^{N_\mathrm{dic}}$. Note that the function $\phi$ before the time-evolution may not be in the spanned subspace $\mathcal{F}_{\bm{\psi}}$, and the function $\phi$ is also approximately written as follows:
\begin{eqnarray}
\phi(\underline{\bm{x}}) \simeq \sum_{i=1}^{N_\mathrm{dic}} c^{\phi}_i \psi_i(\underline{\bm{x}}).
\end{eqnarray}

Since a linear combination of the dictionary functions yields an arbitrary function and its time-evolved function approximately, it is enough for us to consider the action of the Koopman operator on the dictionary. Then, the Koopman operator $\mathcal{K}_{\Delta t_{\mathrm{obs}}}$ is approximated by a finite-size matrix $K_{\Delta t_{\mathrm{obs}}}$ as follows:
\begin{eqnarray}
\left[\begin{array}{c}
\mathcal{K}_{\Delta t_{\mathrm{obs}}} \psi_1(\underline{\bm{x}}) \\ \mathcal{K}_{\Delta t_{\mathrm{obs}}} \psi_2(\underline{\bm{x}}) \\ \vdots \\ \mathcal{K}_{\Delta t_{\mathrm{obs}}} \psi_{N_\mathrm{dic}}(\underline{\bm{x}}) 
\end{array}\right]
=
\mathcal{K}_{\Delta t_{\mathrm{obs}}} \bm{\psi}(\underline{\bm{x}})
\simeq
K_{\Delta t_{\mathrm{obs}}} \bm{\psi}(\underline{\bm{x}}).
\end{eqnarray}
It is easy to determine the Koopman matrix $K_{\Delta t_{\mathrm{obs}}}$ from the dataset; we solve the following least squares problem numerically:
\begin{eqnarray}
K_{\Delta t_{\mathrm{obs}}} = \argmin_{\widetilde{K}_{\Delta t_{\mathrm{obs}}}}  \sum_{n=1}^{N_\mathrm{data}} \left\| \bm{\psi}(\underline{\bm{y}_n}) - \widetilde{K}_{\Delta t_{\mathrm{obs}}} \bm{\psi}(\underline{\bm{x}_n}) \right\|^2.
\label{eq_Koopman_cost_function}
\end{eqnarray}
Note that a comparison with \eref{eq_Kxd_basis_expansion} and \eref{eq_Koopman_cost_function} immediately leads to the element at the $d$-th row and the $i$-th column of the Koopman matrix $K_{\Delta t_{\mathrm{obs}}}$ as follows:
\begin{eqnarray}
\left[K_{\Delta t_{\mathrm{obs}}}\right]_{di} = c^{\mathcal{K}_{\Delta t_{\mathrm{obs}}} x_{d-1}}_i \quad (\textrm{for $2 \le d \le (D+1)$}), \nonumber \\
\left[K_{\Delta t_{\mathrm{obs}}}\right]_{di} = c^{\mathcal{K}_{\Delta t_{\mathrm{obs}}} \psi_d}_i \quad (\textrm{for $d > (D+1)$}).
\end{eqnarray}
Since the first dictionary function $\psi_1(\bm{\underline{x}}) = 1(\bm{\underline{x}})$ is time-invariant, $[K_{\Delta t_{\mathrm{obs}}}]_{1i} = 1$ for all $i$.

In short, one can consider the \textit{linear} action of $\mathcal{K}_{\Delta t_{\mathrm{obs}}}$ in the function space instead of the \textit{nonlinear} time-evolution on the state space, $\bm{F}_{\Delta t_{\mathrm{obs}}}$, in \eref{eq_state_space_evolution}. This linearity naturally leads to the use of a linear combination of basis functions, which facilitates estimation from the data via the least squares method. The usage of linearity is one of the benefits of the Koopman operator approach.

Here, we comment on the relationship between the system's dimensionality and the dictionary size. When we employ monomials in \eref{eq_monomial_dictionary_example} as basis functions, the number of variable combinations increases exponentially as the dimensionality grows. Hence, we require an enormous amount of memory, and the computational cost also becomes high. While there are some studies to avoid this problem (for example, see \cite{Klus2018,Gels2019,Kinjo2025}), the present paper will focus on cases with small dimensionality.

\subsection{Stochastic systems}

As denoted in Sec.~1, the Koopman operator approach is also applicable for stochastic systems; for example, see \cite{Williams2015,Crnjaric-Zic2020,Wanner2022}. In the stochastic cases, the action of the Koopman operator yields an expected value of a statistic for the state variable rather than the state variable itself. For example,
\begin{eqnarray}
\mathcal{K}_{\Delta t_{\mathrm{obs}}} x_1 (\underline{\bm{x}_{\mathrm{ini}}})
= \mathbb{E}[X_1(t+\Delta t_{\mathrm{obs}}) | \underline{\bm{x}(t)} = \underline{\bm{x}_{\mathrm{ini}}}].
\label{eq_action_Koopman_basic}
\end{eqnarray}
In the following, we briefly explain why the Koopman operator approach yields the expectation values from the viewpoint of the Fokker-Planck and the backward Kolmogorov equations.

Consider a $D$-dimensional vector of stochastic variables, $\underline{\bm{X}} \in \mathcal{M}$, and assume that the vector $\underline{\bm{X}(t)}$ at time $t$ obeys the following stochastic differential equation:
\begin{eqnarray}
\rmd \underline{\bm{X}(t)} = \bm{a}(\underline{\bm{X}(t)}) \, \rmd t + B(\underline{\bm{X}(t)}) \, \rmd \underline{\bm{W}(t)},
\label{eq_SDE}
\end{eqnarray}
where $\bm{a}(\underline{\bm{X}})$ is a vector of drift coefficient functions, $B(\underline{\bm{X}})$ is a matrix of diffusion coefficient functions, and $\underline{\bm{W}(t)}$ is a vector of Wiener processes $\{\underline{W_i(t)}\}$. Note that we employ the Ito-type stochastic differential equation  in \eref{eq_SDE}. The Wiener processes satisfy
\begin{eqnarray}
\mathbb{E}[ \underline{W_i(t)} ] = 0, \qquad 
\mathbb{E}[ (\underline{W_i(t)} - \underline{W_i(s)} ) (\underline{W_j(t)} - \underline{W_j(s)} )] = (t-s)\delta_{ij},
\end{eqnarray}
for $0 \le s \le t$. Let $\underline{\bm{X}(0)} = \underline{\bm{x}_\mathrm{ini}}$ be the initial condition for the stochastic differential equation. The stochastic differential equation in \eref{eq_SDE} has a corresponding Fokker-Planck equation \cite{Gardiner_book}, which describes the time-evolution of the probability density function $p(\underline{\bm{x}},t)$:
\begin{eqnarray}
\frac{\partial}{\partial t} p(\underline{\bm{x}},t) = \mathcal{L}^\dagger \, p(\underline{\bm{x}},t),
\label{eq_Fokker_Planck}
\end{eqnarray}
where
\begin{eqnarray}
\mathcal{L}^\dagger
= - \sum_{d} \frac{\partial}{\partial \underline{x_d}} a_d(\underline{\bm{x}}) 
+ \frac{1}{2} \sum_{i,j} \frac{\partial^2}{\partial \underline{x_i} \partial \underline{x_j}} 
\left[ B(\underline{\bm{x}}) B(\underline{\bm{x}})^{\top} \right]_{ij}
\label{eq_L_FP}
\end{eqnarray}
is the time-evolution operator for the Fokker-Planck equation. The initial condition at time $t$ is 
\begin{eqnarray}
p(\underline{\bm{x}}, t) = \delta(\underline{\bm{x}} - \underline{\bm{x}_\mathrm{ini}}),
\end{eqnarray}
where $\delta(\cdot)$ is the Dirac delta function.

Note that the time-evolved probability density function, $p(\underline{\bm{x}}, t+\Delta t_{\mathrm{obs}})$, is formally written as
\begin{eqnarray}
p(\underline{\bm{x}}, t+\Delta t_{\mathrm{obs}}) = \rme^{\mathcal{L}^{\dagger} \Delta t_{\mathrm{obs}}} p(\underline{\bm{x}}, t).
\end{eqnarray}
Then, the expectation for the observable function $x_d$ is given as follows:
\begin{eqnarray}
& \mathbb{E}[X_d(\underline{\bm{x}(t+\Delta t_{\mathrm{obs}})}) | \underline{\bm{x}(t)} = \underline{\bm{x}_{\mathrm{ini}}}] \nonumber \\
&=
\int_{\mathcal{M}} x_d(\underline{\bm{x}}) \, p(\underline{\bm{x}},t+\Delta t_{\mathrm{obs}}) \rmd\underline{\bm{x}} \nonumber \\
&= \int_{\mathcal{M}}
x_d(\underline{\bm{x}})
\left( e^{\mathcal{L}^{\dagger} \Delta t_{\mathrm{obs}}}  \delta(\underline{\bm{x}} - \underline{\bm{x}_\mathrm{ini}}) \right) \rmd\underline{\bm{x}}\nonumber \\
&= \int_{\mathcal{M}}
\left( e^{\mathcal{L} \Delta t_{\mathrm{obs}}}  x_d(\underline{\bm{x}}) \right)
\delta(\underline{\bm{x}} - \underline{\bm{x}_\mathrm{ini}}) \, \rmd\underline{\bm{x}} \nonumber \\
&= \int_{\mathcal{M}} \varphi_{d}(\underline{\bm{x}},t+\Delta t_{\mathrm{obs}}) 
\delta(\underline{\bm{x}} - \underline{\bm{x}_\mathrm{ini}}) \,\rmd\underline{\bm{x}} \nonumber \\
&= \varphi_{d}(\underline{\bm{x}_\mathrm{ini}},t+\Delta t_{\mathrm{obs}}),
\label{eq_duality}
\end{eqnarray}
where $\mathcal{L}$ is the adjoint operator of $\mathcal{L}^\dagger$, i.e.,
\begin{eqnarray}
\mathcal{L}
= \sum_{d} a_d(\bm{x}) \frac{\partial}{\partial x_d} 
+ \frac{1}{2} \sum_{i,j} \left[ \bm{B}(\bm{x}) \bm{B}(\bm{x})^{\top} \right]_{ij} \frac{\partial^2}{\partial x_i \partial x_j},
\label{eq_L}
\end{eqnarray}
and $\varphi_{d}(\underline{\bm{x}},t)$ is the solution of the following partial differential equation, i.e., the so-called the backward Kolmogorov equation:
\begin{eqnarray}
\frac{\partial}{\partial t} \varphi_{d}(\underline{\bm{x}},t) = \mathcal{L} \varphi_{d}(\underline{\bm{x}},t).
\end{eqnarray}
The backward Kolmogorov equation leads to the following formal solution:
\begin{eqnarray}
\varphi_{d}(\underline{\bm{x}},t+\Delta t_{\mathrm{obs}}) = \rme^{\mathcal{L} \Delta t_{\mathrm{obs}}} \varphi_{d}(\underline{\bm{x}},t),
\label{eq_Koopman_pre}
\end{eqnarray}
where $\varphi_{d}(\underline{\bm{x}},t+\Delta t_{\mathrm{obs}}) \in \mathcal{F}$, and the initial condition is 
\begin{eqnarray}
\varphi_{d}(\underline{\bm{x}},t) = x_d(\underline{\bm{x}}).
\end{eqnarray}

Here, $\mathcal{L}^\dagger$ in \eref{eq_L_FP} yields the time-evolution of the probability density function on the state space $\mathcal{M}$. By contrast, $\mathcal{L}$ in \eref{eq_L} corresponds to the time-evolution of functions on the function space $\mathcal{F}$. In addition, \eref{eq_Koopman_pre} indicates that 
\begin{eqnarray}
\mathcal{K} = \rme^{\mathcal{L} \Delta t_{\mathrm{obs}}},
\end{eqnarray}
and it is straightforwardly understandable that the Koopman operator $\mathcal{K}$ yields the expectation after the time-evolution.

Note that the above discussion is also available to the deterministic cases. When there is no diffusion, i.e., $B(\underline{\bm{x}}) = 0$, the time-evolution operator $\mathcal{L}^\dagger$ includes only the first-order derivatives. While the second-order derivatives lead to the effect of broadening the probability density function, the first-order derivatives correspond to the shift of the coordinates. Then, the zero diffusion cases retain the probability density function as the Dirac delta function; $\delta(\underline{\bm{x}} - \underline{\bm{x}_\mathrm{ini}}) \to \delta(\underline{\bm{x}} - \underline{\bm{x}(t+\Delta t_{\mathrm{obs}})})$. Hence, the expectation yields the value of $\underline{\bm{x}(t+\Delta t_{\mathrm{obs}})}$, and the previous discussions on the deterministic cases are recovered adequately. 

There are some comments on the choice of $\Delta t_{\mathrm{obs}}$. As discussed above, we approximate the Koopman operator $\mathcal{K}$ with a corresponding matrix $K$ in terms of basis functions, i.e., a dictionary. If one employs longer observation intervals $\Delta t_{\mathrm{obs}}$, more basis functions will be required, and hence, there is a considerable increase in computational cost. In addition, it is well-known that significantly long observation intervals require exponentially large basis functions, making the use of the Koopman operator impractical. Hence, in the numerical experiments in Sec.~5, we will employ reasonable observation intervals. Note that slight changes in the observation interval do not affect the discussions and conclusions in the present paper.

\section{Revisit on partial observations in deterministic systems}

\subsection{Use of time-dependent basis}

Here, we follow \cite{LinYT2021} and revisit the effects of partial observations in deterministic systems. Note that the following discussions are slightly different from \cite{LinYT2021}. As discussed below, the absence of diffusion coefficients $B(\bm{x})$ is crucial to connect the Koopman operator approach and the conventional discussions for partial observations, i.e., the Mori-Zwanzig formalism.

Assume that only some of the coordinates are observable, and we denote the index set as $\mathcal{O}$. The vector of partial observables is $\bm{g}_{\mathcal{O}}$ whose elements are $\{x_d | d \in \mathcal{O}\}$. The number of partial observables is $N_{\mathrm{obs}}$. For example, when we observe $\underline{x_1}$ and $\underline{x_3}$ in a $D=3$ case, $\mathcal{O} = \{1,3\}$, $\bm{g}_{\mathcal{O}} = [x_1 \,\, x_3]^{\top}$, and $N_{\mathrm{obs}} = 2$. We also introduce a dictionary $\bm{\psi}_{\overline{\mathcal{O}}}$ that includes many basis functions $\psi_{i} \in \mathcal{F}$ related to the unobserved variables. We denote the number of dictionary functions as $N_{\mathrm{dic}}^{\overline{\mathcal{O}}}$. The discussions on the function space and the introduction of the basis functions, i.e., the dictionary, lead to a linear algebraic approach for the nonlinear systems; this linearity is characteristic of the Koopman operator approach.

We discuss a case where the basis functions are time-dependent, which corresponds to the discussion in \cite{LinYT2021}. In this case, an arbitrary function $\phi \in \mathcal{F}$ is expressed approximately as
\begin{eqnarray}
\phi(\underline{\bm{x}},t) = \sum_{d \in \mathcal{O}} c^{\phi,\mathcal{O}}_d x_d(\underline{\bm{x}},t)
+ \sum_{i = 1}^{N_\mathrm{dic}^{\overline{\mathcal{O}}}} c^{\phi,\overline{\mathcal{O}}}_i \psi_i(\underline{\bm{x}},t).
\label{eq_phi_expansion}
\end{eqnarray}
Then, we derive an explicit representation of the time-evolution operator $\mathcal{L}$ in \eref{eq_L}. Using the same basis functions above, we approximate the operator $\mathcal{L}$ as matrices; it is enough to consider the actions on $\{x_d\}$ and $\{\psi_i\}$, and we have
\begin{eqnarray}
\mathcal{L} x_d(\underline{\bm{x}},t) \simeq \sum_{d' \in \mathcal{O}} L^{\mathcal{O} \mathcal{O}}_{d d'} x_{d'}(\underline{\bm{x}},t)
+ \sum_{i = 1}^{N_\mathrm{dic}^{\overline{\mathcal{O}}}} L^{\mathcal{O} \overline{\mathcal{O}}}_{d i} \psi_i(\underline{\bm{x}},t),
\end{eqnarray}
and
\begin{eqnarray}
\mathcal{L} \psi_i(\underline{\bm{x}},t) \simeq \sum_{d \in \mathcal{O}} L^{\overline{\mathcal{O}} \mathcal{O}}_{i d} x_{d}(\underline{\bm{x}},t)
+ \sum_{i' = 1}^{N_\mathrm{dic}^{\overline{\mathcal{O}}}} L^{\overline{\mathcal{O}} \overline{\mathcal{O}}}_{i i'} \psi_{i'}(\underline{\bm{x}},t).
\end{eqnarray}
Then, using the matrices, $L^{\mathcal{O} \mathcal{O}} \in \mathbb{R}^{N_{\mathrm{obs}} \times N_{\mathrm{obs}}}$, $L^{\mathcal{O} \overline{\mathcal{O}}} \in \mathbb{R}^{N_{\mathrm{obs}} \times N_{\mathrm{dic}}^{\overline{\mathcal{O}}}}$, $L^{\overline{\mathcal{O}} \mathcal{O}} \in \mathbb{R}^{N_{\mathrm{dic}}^{\overline{\mathcal{O}}} \times N_{\mathrm{obs}}}$ and $L^{\overline{\mathcal{O}} \overline{\mathcal{O}}} \in \mathbb{R}^{N_{\mathrm{dic}}^{\overline{\mathcal{O}}} \times N_{\mathrm{dic}}^{\overline{\mathcal{O}}}}$, the time-evolution equation is given as
\begin{eqnarray}
\frac{\rmd}{\rmd t} 
\left[
\begin{array}{c}
\bm{g}_{\mathcal{O}}(t) \\
\bm{\psi}_{\overline{\mathcal{O}}}(t)
\end{array}
\right]
= 
\left[
\begin{array}{cc}
L^{\mathcal{O} \mathcal{O}} & L^{\mathcal{O} \overline{\mathcal{O}}}\\
L^{\overline{\mathcal{O}} \mathcal{O}} & L^{\overline{\mathcal{O}} \overline{\mathcal{O}}}
\end{array}
\right]
\left[
\begin{array}{c}
\bm{g}_{\mathcal{O}}(t) \\
\bm{\psi}_{\overline{\mathcal{O}}}(t)
\end{array}
\right].
\label{eq_time_dependent_MZ_1}
\end{eqnarray}
As shown in \cite{LinYT2021}, it is possible to derive the time-evolution equation for $\bm{g}_{\mathcal{O}}(t)$. First, we solve \eref{eq_time_dependent_MZ_1} for $\bm{\psi}_{\overline{\mathcal{O}}}(t)$ implicitly;
\begin{eqnarray}
\bm{\psi}_{\overline{\mathcal{O}}}(t) 
= 
\int_0^{t} \rme^{(t-s) L^{\overline{\mathcal{O}} \overline{\mathcal{O}}}} L^{\overline{\mathcal{O}} \mathcal{O}} \bm{g}_{\mathcal{O}}(s) \rmd s
+ \rme^{t L^{\overline{\mathcal{O}} \overline{\mathcal{O}}}} \bm{\psi}_{\overline{\mathcal{O}}}(0).
\label{eq_time_dependent_MZ_2}
\end{eqnarray}
Then, inserting \eref{eq_time_dependent_MZ_2} to \eref{eq_time_dependent_MZ_1}, we have
\begin{eqnarray}
\fl
\frac{\rmd}{\rmd t} \bm{g}_{\mathcal{O}}(t)
=
L^{\mathcal{O} \mathcal{O}} \bm{g}_{\mathcal{O}}(t)
+ L^{\mathcal{O} \overline{\mathcal{O}}} 
\int_0^{t} \rme^{(t-s) L^{\overline{\mathcal{O}} \overline{\mathcal{O}}}} L^{\overline{\mathcal{O}} \mathcal{O}} \bm{g}_{\mathcal{O}}(s) \rmd s
+  L^{\mathcal{O} \overline{\mathcal{O}}} \rme^{t L^{\overline{\mathcal{O}} \overline{\mathcal{O}}}} \bm{\psi}_{\overline{\mathcal{O}}}(0).
\label{eq_time_dependent_MZ_3}
\end{eqnarray}
Replacing $L^{\mathcal{O} \mathcal{O}}$ with $M_{\mathrm{trans}}$ and $- L^{\mathcal{O} \overline{\mathcal{O}}} \rme^{s L^{\overline{\mathcal{O}} \overline{\mathcal{O}}}} L^{\overline{\mathcal{O}} \mathcal{O}}$ with $K_{\mathrm{mem}}(s)$, we have the following generalized Langevin equation:
\begin{eqnarray}
\frac{\rmd}{\rmd t} \bm{g}_{\mathcal{O}}(t)
=
M_{\mathrm{trans}} \, \bm{g}_{\mathcal{O}}(t)
- \int_0^{t} K_{\mathrm{mem}}(t-s) \bm{g}_{\mathcal{O}}(s) \rmd s + \bm{f}(t),
\label{eq_time_dependent_MZ_final}
\end{eqnarray}
where $\bm{f}(t) = L^{\mathcal{O} \overline{\mathcal{O}}} \rme^{t L^{\overline{\mathcal{O}} \overline{\mathcal{O}}}} \bm{\psi}_{\overline{\mathcal{O}}}(0)$. $M_{\mathrm{trans}}$ is called the Markov transition matrix, which quantifies the interactions between the observables $\bm{g}_{\mathcal{O}}(t)$. $K_{\mathrm{mem}}(s)$ is called the memory kernel and depends on the history of the observed quantities; \eref{eq_time_dependent_MZ_3} indicates that the memory effect stems from the path via the unobserved quantities. $\bm{f}(t)$ is referred to as noise since it depends on the initial values of the unobserved quantities $\bm{\psi}_{\overline{\mathcal{O}}}(0)$. That is, there is no information about their explicit values, and then we cannot predict them.

\subsection{Feature of deterministic systems}

Note that \eref{eq_time_dependent_MZ_final} is an equation in the function space $\mathcal{F}$. It is crucial to distinguish the function space $\mathcal{F}$ and the state space $\mathcal{M}$; an element $x_d$ in $\bm{g}_{\mathcal{O}}(t)$ is different from an element $\underline{x_d}$ in the state vector $\underline{\bm{x}}$ in general. However, deterministic systems have a unique feature that one can interpret \eref{eq_time_dependent_MZ_final} as an equation for the state vector.

First, we focus on the time-evolution equation in \eref{eq_duality}; we see that the final expression $\varphi_{d}(\underline{\bm{x}_\mathrm{ini}},t+\Delta t_{\mathrm{obs}})$ yields the expectation values. Although the expectation value $\mathbb{E}[x_d]$ is not the state $\underline{x_d}$, the absence of the noise term does not change an initial Dirac-delta-type density function even in the time-evolution, as discussed in Sec.~2.4. That is,
\begin{eqnarray}
e^{\mathcal{L}^{\dagger} \Delta t_{\mathrm{obs}}}  \delta(\underline{\bm{x}} - \underline{\bm{x}(t)})
= \delta(\underline{\bm{x}} - \underline{\bm{x}(t+\Delta t_{\mathrm{obs}})}).
\end{eqnarray}
Then, the expectation value is equal to the corresponding state;
\begin{eqnarray}
\mathbb{E}[X_d(\underline{\bm{x}(t+\Delta t_{\mathrm{obs}})}) | \underline{\bm{x}(t)} = \underline{\bm{x}_{\mathrm{ini}}}]
= \underline{x_d(t+\Delta t_{\mathrm{obs}})}.
\end{eqnarray}
This fact enables us to equate the observable function $x_d$ with the state $\underline{x_d}$.

Second, we introduce a different perspective on the time-evolution equation. In deterministic systems,  there is no noise, i.e., $B(\bm{x}) = 0$. Hence, the second term on the right-hand side of \eref{eq_L} is absent. Then, the following operator yields the time-evolution in the function space:
\begin{eqnarray}
\mathcal{L} = \sum_{d} a_d(\bm{x}) \frac{\partial}{\partial x_d}.
\end{eqnarray}
When we consider the time-evolution for the observable function $x_d$, the action of $\mathcal{L}$ leads to
\begin{eqnarray}
\mathcal{L}x_d = \sum_{d'} a_{d'}(\bm{x}) \frac{\partial}{\partial x_{d'}} x_d = a_d(\bm{x}),
\end{eqnarray}
and then, we have 
\begin{eqnarray}
\frac{\rmd}{\rmd t} x_d = a_d(\bm{x}).
\label{eq_deterministic_derived_ode}
\end{eqnarray}
Note that \eref{eq_deterministic_derived_ode} is the same form as the original time-evolution equation in the state space:
\begin{eqnarray}
\frac{\rmd}{\rmd t} \underline{x_d} = a_d(\underline{\bm{x}}).
\end{eqnarray}
This fact also allows us to equate the time-evolution of $x_d$ in the function space $\mathcal{F}$ with that of $\underline{x_d}$ in the state space $\mathcal{M}$.

In short, although we should distinguish the function space and the state space, the feature of the deterministic system enables us to consider the derived equation \eref{eq_time_dependent_MZ_final} in the function space as the one in the state space. This feature justifies the coarse-grained approach based on the Mori-Zwanzig formalism; we can construct time-evolution equations of a smaller set of macroscopic variables that are measurable, and other unmeasured degrees of freedom play the role of noise.

\section{Partial observations in stochastic systems}

\subsection{Discussion with time-independent basis functions}

As discussed in Sec.~3, the derived equation in \eref{eq_time_dependent_MZ_final} is defined in the function space. Note that the function at time $t+\Delta t_{\mathrm{obs}}$ yields the expectation values in \eref{eq_duality}. Due to the presence of the noise term, even if the initial density function is the Dirac delta function, the values of $\mathbb{E}[x_d(\underline{\bm{x}(t+\Delta t_{\mathrm{obs}})})]$ and $\underline{x_d}(t+\Delta t_{\mathrm{obs}})$ are different because the density function is spread out. Hence, the generalized Langevin equation in \eref{eq_time_dependent_MZ_final} is not directly related to the state vector $\underline{\bm{x}}$. 

Here, we discuss the generalized Langevin equation in the function space. Although the time-dependent basis in Sec.~3 is beneficial due to the connection with the state vector, it would be preferable to fix the basis functions in the discussion below. Then, we employ the following basis expansion:
\begin{eqnarray}
\phi(\underline{\bm{x}},t) = \sum_{d \in \mathcal{O}} c^{\phi,\mathcal{O}}_d(t) x_d(\underline{\bm{x}})
+ \sum_{i = 1}^{N_{\mathrm{dic}}^{\overline{\mathcal{O}}}} c^{\phi,\overline{\mathcal{O}}}_i(t)  \psi_i(\underline{\bm{x}}),
\label{eq_phi_expansion_stochastic}
\end{eqnarray}
where the expansion coefficients $\{c^{\phi,\mathcal{O}}_d(t)\}$ and $\{c^{\phi,\overline{\mathcal{O}}}_i(t)\}$ are time-dependent, which differs from \eref{eq_phi_expansion}.

Then, we derive the time-evolution equation for the expansion coefficients $\{c^{\phi,\mathcal{O}}_d(t)\}$ and $\{c^{\phi,\overline{\mathcal{O}}}_i(t)\}$. In the derivation, we employ the bra and ket notations for notational simplicity. Note that the notations are essentially the same as the Doi-Peliti method \cite{Doi1976,Doi1976a,Peliti1985}; see the discussion in \cite{Ohkubo2013,Ohkubo2019,Takahashi2023}.

In the function space, we introduce the following ket states $\{|x_d \rangle\}$ and $\{|\psi_i\rangle\}$:
\begin{eqnarray}
|x_d \rangle = x_d \qquad \textrm{for } d = 1, \dots, D, \label{eq_ket_1}\\
|\psi_i\rangle = \psi_i \qquad \textrm{for } i = 1, \dots, N_{\mathrm{dic}}^{\overline{\mathcal{O}}}, \label{eq_ket_2}
\end{eqnarray}
where we omit the argument of functions, $\underline{\bm{x}}$. By contrast, the bra states $\{\langle x_d |\}$ and $\{\langle \psi_i |\}$ are defined as follows:
\begin{eqnarray}
\langle x_d | = \int \prod_{d'=1}^D \rmd x_{d'} \,\delta(x_{d'}) \left( \frac{\partial}{\partial x_d}\right) (\cdot), \label{eq_bra_1}\\
\langle \psi_i | =
\int \rmd\bm{x} \,\rho(\bm{x}) \psi_i(\bm{x}) (\cdot), \label{eq_bra_2}
\end{eqnarray}
where $\rho(\bm{x})$ is a suitable measure to yield $L^2$ functions. From \eref{eq_ket_1} and \eref{eq_bra_1}, we have the following orthonormal relation:
\begin{eqnarray}
\langle x_{d} | x_{d'} \rangle = \int \rmd x_d \,\delta(x_d) \left( \frac{\partial}{\partial x_d}\right) x_{d'} = \delta_{d,d'},
\label{eq_naive_orthogonal}
\end{eqnarray}
where $\delta_{d,d'}$ is the Kronecker delta function.

It would be easy to explain with a concrete example. Hence, we here assume that there are only two observed variables, $\mathcal{O} = \{1,2\}$. Using the above bra and ket notations, \eref{eq_phi_expansion_stochastic} is rewritten as
\begin{eqnarray}
\phi(\underline{\bm{x}},t) = \sum_{d \in \mathcal{O}} c^{\phi,\mathcal{O}}_d(t) | x_d \rangle
+ \sum_{i = 1}^{N_{\mathrm{dic}}^{\overline{\mathcal{O}}}} c^{\phi,\overline{\mathcal{O}}}_i(t)  | \psi_i \rangle.
\label{eq_phi_expansion_stochastic_2}
\end{eqnarray}
The time-derivative of \eref{eq_phi_expansion_stochastic_2} leads to 
\begin{eqnarray}
\frac{\rmd}{\rmd t} \phi(\underline{\bm{x}},t) = \sum_{d \in \mathcal{O}} \frac{\rmd}{\rmd t} \left(c^{\phi,\mathcal{O}}_d(t) \right) | x_d \rangle
+ \sum_{i = 1}^{N_{\mathrm{dic}}^{\overline{\mathcal{O}}}} \frac{\rmd}{\rmd t} \left( c^{\phi,\overline{\mathcal{O}}}_i(t)  \right) | \psi_i \rangle.
\label{eq_phi_expansion_stochastic_dt}
\end{eqnarray}
Next, we take the action of $\langle x_d|$ on all terms from the left. For example, the left action of $\langle x_1|$ on \eref{eq_phi_expansion_stochastic_dt} yields
\begin{eqnarray}
\fl
\langle x_1| \frac{\rmd}{\rmd t} \phi(\underline{\bm{x}},t)
= \sum_{d \in \mathcal{O}} \frac{\rmd}{\rmd t} \left(c^{\phi,\mathcal{O}}_d(t) \right) \langle x_1|  x_d \rangle
+ \sum_{i = 1}^{N_{\mathrm{dic}}^{\overline{\mathcal{O}}}} \frac{\rmd}{\rmd t} \left( c^{\phi,\overline{\mathcal{O}}}_i(t)  \right) \langle x_1| \psi_i \rangle \nonumber \\
= \frac{\rmd}{\rmd t} \left(c^{\phi,\mathcal{O}}_1(t) \right) \langle x_1|  x_1 \rangle
+ \sum_{i = 1}^{N_{\mathrm{dic}}^{\overline{\mathcal{O}}}} \frac{\rmd}{\rmd t} \left( c^{\phi,\overline{\mathcal{O}}}_i(t)  \right) \langle x_1| \psi_i \rangle \nonumber \\
= 
\left[
\begin{array}{ccccc}
1 & 0 & \langle x_1| \psi_1 \rangle & \cdots &  \langle x_1 | \psi_{N_{\mathrm{dic}}^{\overline{\mathcal{O}}}} \rangle
\end{array}
\right]
\left[
\begin{array}{c}
c^{\phi,\mathcal{O}}_1(t) \\
c^{\phi,\mathcal{O}}_2(t) \\
c^{\phi,\overline{\mathcal{O}}}_1(t) \\
\vdots\\
c^{\phi,\overline{\mathcal{O}}}_{N_{\mathrm{dic}}^{\overline{\mathcal{O}}}}(t) 
\end{array}
\right],
\end{eqnarray}
where we used $\langle x_d|  x_{d'} \rangle = \delta_{d,d'}$ in \eref{eq_naive_orthogonal}. The left action of $\langle x_1|$ on $\mathcal{L} \phi(\underline{\bm{x}},t)$ gives
\begin{eqnarray}
\fl
\langle x_1| \mathcal{L} \phi(\underline{\bm{x}},t)
= 
\sum_{d \in \mathcal{O}} \left(c^{\phi,\mathcal{O}}_d(t) \right) \langle x_1| \mathcal{L} |  x_d \rangle
+ \sum_{i = 1}^{N_{\mathrm{dic}}^{\overline{\mathcal{O}}}} \left( c^{\phi,\overline{\mathcal{O}}}_i(t)  \right) \langle x_1| \mathcal{L} |\psi_i \rangle \nonumber \\
= 
\left[
\begin{array}{ccccc}
\widetilde{L}^{\mathcal{O} \mathcal{O}}_{11}  & \widetilde{L}^{\mathcal{O} \mathcal{O}}_{12} & 
\widetilde{L}^{\mathcal{O} \overline{\mathcal{O}}}_{1 1} &
\cdots & 
\widetilde{L}^{\mathcal{O} \overline{\mathcal{O}}}_{1 N_{\mathrm{dic}}^{\overline{\mathcal{O}}}}
\end{array}
\right]
\left[
\begin{array}{c}
c^{\phi,\mathcal{O}}_1(t) \\
c^{\phi,\mathcal{O}}_2(t) \\
c^{\phi,\overline{\mathcal{O}}}_1(t) \\
\vdots\\
c^{\phi,\overline{\mathcal{O}}}_{N_{\mathrm{dic}}^{\overline{\mathcal{O}}}}(t) 
\end{array}
\right],
\end{eqnarray}
where 
\begin{eqnarray}
\langle x_d | \mathcal{L} | x_{d'} \rangle = \widetilde{L}^{\mathcal{O} \mathcal{O}}_{d d'}, 
\quad \langle x_d | \mathcal{L} | \psi_i \rangle = \widetilde{L}^{\mathcal{O} \overline{\mathcal{O}}}_{d i}.
\end{eqnarray}
Similarly, we calculate the left actions of $\langle x_2|$ and $\{\langle \psi_i|\}$ and combine these results. Then, we have the following time-evolution equation:
\begin{eqnarray}
Z
\frac{\rmd}{\rmd t}
\left[
\begin{array}{c}
\bm{c}^{\phi,\mathcal{O}}(t)\\
\bm{c}^{\phi,\overline{\mathcal{O}}}(t)\\
\end{array}
\right]
= \left[
\begin{array}{cc}
\widetilde{L}^{\mathcal{O} \mathcal{O}} & \widetilde{L}^{\mathcal{O} \overline{\mathcal{O}}}\\
\widetilde{L}^{\overline{\mathcal{O}} \mathcal{O}} & \widetilde{L}^{\overline{\mathcal{O}} \overline{\mathcal{O}}}
\end{array}
\right]
\left[
\begin{array}{c}
\bm{c}^{\phi,\mathcal{O}}(t)\\
\bm{c}^{\phi,\overline{\mathcal{O}}}(t)\\
\end{array}
\right],
\label{eq_c_time_evolution_pre}
\end{eqnarray}
where we define
\begin{eqnarray}
\fl
Z =
\left[
\begin{array}{cccccc}
1 & 0 & \langle x_1 | \psi_1 \rangle & \langle x_1 | \psi_2 \rangle & \cdots & \langle x_1 | \psi_{N_{\mathrm{dic}}^{\overline{\mathcal{O}}}} \rangle \\
0 & 1 & \langle x_2 | \psi_1 \rangle & \langle x_2 | \psi_2 \rangle & \cdots & \langle x_2 | \psi_{N_{\mathrm{dic}}^{\overline{\mathcal{O}}}} \rangle \\
\langle \psi_1 | x_1 \rangle & \langle \psi_1 | x_2 \rangle  
& \langle \psi_1 | \psi_1 \rangle & \langle \psi_1 | \psi_2 \rangle & \cdots & \langle \psi_1 | \psi_{N_{\mathrm{dic}}^{\overline{\mathcal{O}}}} \rangle \\
\vdots & \vdots & \vdots & \vdots & \ddots & \vdots \\
\langle \psi_{N_{\mathrm{dic}}^{\overline{\mathcal{O}}}}  | x_1 \rangle & \langle \psi_{N_{\mathrm{dic}}^{\overline{\mathcal{O}}}}  | x_2 \rangle  
& \langle \psi_{N_{\mathrm{dic}}^{\overline{\mathcal{O}}}}  | \psi_1 \rangle & \langle \psi_{N_{\mathrm{dic}}^{\overline{\mathcal{O}}}}  | \psi_2 \rangle & \cdots & \langle \psi_{N_{\mathrm{dic}}^{\overline{\mathcal{O}}}}  | \psi_{N_{\mathrm{dic}}^{\overline{\mathcal{O}}}} \rangle 
\end{array}
\right]
\label{eq_Z_pre}
\end{eqnarray}
and
\begin{eqnarray}
\quad \langle \psi_i | \mathcal{L} | x_{d'} \rangle = \widetilde{L}^{\overline{\mathcal{O}} \mathcal{O}}_{i d'}, 
\quad \langle \psi_i | \mathcal{L} | \psi_{i'} \rangle = \widetilde{L}^{\overline{\mathcal{O}} \overline{\mathcal{O}}}_{i i'}.
\end{eqnarray}

Note that the matrix $Z$ in \eref{eq_Z_pre} is not diagonal. Hence, we need some effort to evaluate the inverse matrix $Z^{-1}$. However, the following assumption for the dictionary functions $\{|\psi_i\rangle\}$ makes the discussion simple: The functions $\{|\psi_i\rangle\}$ consist of monomial basis functions. Then, an element in $\{|\psi_i\rangle\}$ has the following form:
\begin{eqnarray}
|\bm{n}\rangle \equiv \prod_{d=1}^D x_d^{n_d},
\end{eqnarray}
where $\bm{n} = \{n_1,n_2,\dots,n_D\}$. The corresponding bra state is defined as
\begin{eqnarray}
\langle \bm{n} | \equiv \int \prod_{d=1}^D \rmd x_d \,\delta(x_d) \left( \frac{\partial}{\partial x_d}\right)^{n_d} (\cdot).
\end{eqnarray}
Hence, we have the following orthogonal relation:
\begin{eqnarray}
\langle \bm{n} | \bm{m} \rangle = \int \prod_{d=1}^D \rmd x_d \,\delta(x_d) \left( \frac{\partial}{\partial x_d}\right)^{n_d} x_{d}^{m_d} 
= \prod_{d=1}^D n_d! \delta_{n_d, m_d}.
\end{eqnarray}
Note that the monomial functions for the observed states,
\begin{eqnarray}
| \{1,0,\dots,0\} \rangle = | x_1 \rangle \quad \textrm{and} \quad 
| \{0,1,\dots,0\} \rangle = | x_2 \rangle,
\end{eqnarray}
should not be included in $\{|\psi_i\rangle\}$. Using the above assumption for the dictionary functions, we have the following matrix $Z$ in \eref{eq_Z_pre}:
\begin{eqnarray}
Z =
\left[
\begin{array}{cccccc}
1 & 0 & 0   & 0 & \cdots & 0 \\
0 & 1 & 0   & 0 & \cdots & 0\\
0 & 0 & z_1 & 0 & \cdots & 0 \\
\vdots & \vdots & \vdots & \vdots & \ddots & \vdots \\
0 & 0 & 0   & 0 & \cdots & z_{N_{\mathrm{dic}}^{\overline{\mathcal{O}}}}
\end{array}
\right],
\end{eqnarray}
where we denote the diagonal part with $\{z_i | i = 1, \dots, N_{\mathrm{dic}}^{\overline{\mathcal{O}}}\}$. Since the matrix $Z$ is diagonal, $Z^{-1}$ is also diagonal, which leads to
\begin{eqnarray}
\frac{\rmd}{\rmd t} 
\left[
\begin{array}{c}
\bm{c}^{\phi,\mathcal{O}}(t)\\
\bm{c}^{\phi,\overline{\mathcal{O}}}(t)\\
\end{array}
\right]
&= 
Z^{-1}
\left[
\begin{array}{cc}
\widetilde{L}^{\mathcal{O} \mathcal{O}} & \widetilde{L}^{\mathcal{O} \overline{\mathcal{O}}}\\
\widetilde{L}^{\overline{\mathcal{O}} \mathcal{O}} & \widetilde{L}^{\overline{\mathcal{O}} \overline{\mathcal{O}}}
\end{array}
\right]
\left[
\begin{array}{c}
\bm{c}^{\phi,\mathcal{O}}(t)\\
\bm{c}^{\phi,\overline{\mathcal{O}}}(t)\\
\end{array}
\right]\nonumber \\
&=
\left[
\begin{array}{cc}
\widetilde{L}^{\mathcal{O} \mathcal{O}} & \widetilde{L'}^{\mathcal{O} \overline{\mathcal{O}}}\\
\widetilde{L'}^{\overline{\mathcal{O}} \mathcal{O}} & \widetilde{L'}^{\overline{\mathcal{O}} \overline{\mathcal{O}}}
\end{array}
\right]
\left[
\begin{array}{c}
\bm{c}^{\phi,\mathcal{O}}(t)\\
\bm{c}^{\phi,\overline{\mathcal{O}}}(t)\\
\end{array}
\right].
\label{eq_c_time_evolution}
\end{eqnarray}
Note that the multiplication of the diagonal matrix $Z^{-1}$ does not change the relationship between observed and unobserved functions. Since \eref{eq_c_time_evolution} has the same form as \eref{eq_time_dependent_MZ_1}, we finally obtain the following equation similar to \eref{eq_time_dependent_MZ_final}:
\begin{eqnarray}
\fl
\frac{\rmd}{\rmd t} \bm{c}^{\phi,\mathcal{O}}(t)
=
\widetilde{L}^{\mathcal{O} \mathcal{O}} \bm{c}^{\phi,\mathcal{O}}(t)
+ \widetilde{L'}^{\mathcal{O} \overline{\mathcal{O}}} 
\int_0^{t} \rme^{(t-s) \widetilde{L'}^{\overline{\mathcal{O}} \overline{\mathcal{O}}}} \widetilde{L'}^{\overline{\mathcal{O}} \mathcal{O}} \bm{c}^{\phi,\mathcal{O}}(s) \rmd s
+  \widetilde{L'}^{\mathcal{O} \overline{\mathcal{O}}} \rme^{t \widetilde{L'}^{\overline{\mathcal{O}} \overline{\mathcal{O}}}} \bm{c}^{\phi,\overline{\mathcal{O}}}(0).
\label{eq_c_MZ_final_1}
\end{eqnarray}

Although the form of \eref{eq_c_MZ_final_1} is similar to \eref{eq_time_dependent_MZ_1}, there are differences. First, \eref{eq_c_MZ_final_1} is a vector-state equation for only one function $\phi$. Second, the initial values of the coefficients $\{\bm{c}^{\phi,\mathcal{O}}(0)\}$ are one-hot, i.e., all elements are zero except for a single one at a unique position, because we focus only on the observable functions $\{x_d | d \in \mathcal{O}\}$. For example, assume that we focus on $x_1$. Hence, we have the following equation for $\phi = x_1$ at the initial time $t=0$:
\begin{eqnarray}
x_1 = \phi(\underline{\bm{x}},0) = \sum_{d \in \mathcal{O}} c^{\phi,\mathcal{O}}_d(0) | x_d \rangle
+ \sum_{i = 1}^{N_{\mathrm{dic}}^{\overline{\mathcal{O}}}} c^{\phi,\overline{\mathcal{O}}}_i(0)  | \psi_i \rangle,
\end{eqnarray}
which indicates that only the first element, $c^{\phi,\mathcal{O}}_1(0)$, is 1, and the other terms become zero because the left-hand side is the function $x_1$. Then, different from the time-dependent basis functions, the third term in the right-hand side in \eref{eq_c_MZ_final_1}, i.e., the noise term, vanishes;
\begin{eqnarray}
\frac{\rmd}{\rmd t} \bm{c}^{\phi,\mathcal{O}}(t)
=
\widetilde{M}_{\mathrm{trans}} \, \bm{c}^{\phi,\mathcal{O}}(t)
- \int_0^{t} \widetilde{K}_{\mathrm{mem}}(t-s) \bm{c}^{\phi,\mathcal{O}}(s) \rmd s,
\label{eq_c_MZ_final_2}
\end{eqnarray}
where $\widetilde{M}_{\mathrm{trans}} \equiv \widetilde{L}^{\mathcal{O} \mathcal{O}}$ and $\widetilde{K}_{\mathrm{mem}}(s) \equiv - \widetilde{L'}^{\mathcal{O} \overline{\mathcal{O}}} \rme^{s \widetilde{L'}^{\overline{\mathcal{O}} \overline{\mathcal{O}}}} \widetilde{L'}^{\overline{\mathcal{O}} \mathcal{O}}$. Note that, different from \eref{eq_time_dependent_MZ_final}, it is not straightforwardly clear that the use of time-delayed states is beneficial. The second term on the right-hand side in \eref{eq_c_MZ_final_2} is related to the expansion coefficient, which does not directly correspond to the delayed state.

\subsection{Effectiveness of delay-embedding}

As mentioned above, it is crucial to distinguish the state and function spaces. The discussions in Sec.~3.1 and Sec.~4.1 are based on the Koopman operator approach which focuses on the function space. The discussion in Sec.~3.2 focuses on the feature of the deterministic systems, which yields the connection with the state and function spaces. 

It is well-known that delay-embedding is effective in deterministic systems \cite{Arbabi2017,Brunton2017,Wanner2022,Clainche2017,Kamb2020}. The effectiveness of delay-embedding could be understandable from the viewpoint of the Takens theorem \cite{Takens1981}; see also the discussions in \cite{LinYT2021}. However, in stochastic systems, it is not straightforward to apply the discussion based on the Takens theorem because we consider the function space. Here, we discuss the effectiveness of the delay-embedding in the stochastic systems.

As discussed in Sec.~2, the introduction of the dictionary enables us to deal with the nonlinearity within the Koopman operator approach. However, assuming the monomial basis functions, the use of higher-order basis functions causes the so-called curse of dimensionality, which exponentially increases the size of the dictionary. In the discussion in Sec.~4.1, the basis functions affect the second term on the right-hand sides in \eref{eq_c_MZ_final_1} and \eref{eq_c_MZ_final_2}. The approximation by a finite dictionary can significantly reduce accuracy.

Here, note that the Koopman operator approach leads to 
\begin{eqnarray}
x_d(t) = \mathcal{K}^{\Delta t_{\mathrm{obs}}} x_d(t-\Delta t_{\mathrm{obs}}) = \rme^{\mathcal{L}\Delta t_{\mathrm{obs}}} x_d (t-\Delta t_{\mathrm{obs}})
\label{eq_delay_explanation_1}
\end{eqnarray}
for an observable function $x_d$. Then, we rewrite \eref{eq_delay_explanation_1} formally as follows:
\begin{eqnarray}
\fl
x_d (t-\Delta t_{\mathrm{obs}}) \nonumber \\
\fl
= \rme^{- \mathcal{L}\Delta t_{\mathrm{obs}}} x_d(t)  \nonumber \\
\fl
= 
\sum_{k=0}^\infty \left(
- \sum_{d'}^N a_{d'}(\bm{x}) \frac{\partial}{\partial x_{d'}}
- \frac{1}{2} \sum_{i,j} [B(\bm{x})B(\bm{x})]^{\top}_{i,j} \frac{\partial^2}{\partial x_{i} \partial x_{j}}
\right)^k
(\Delta t_{\mathrm{obs}})^k
x_d(t).
\label{eq_delay_explanation_2}
\end{eqnarray}
Since $a_d(\bm{x})$ and $B(\bm{x})$ include various functions, the exponential in \eref{eq_delay_explanation_2} yields various basis functions not included in the finite dictionary. Hence, they are suitable for representing the subspace spanned by the time-evolution operator, and we could expect that the delay-embedding will work well. In the next section, we will demonstrate the effectiveness of the delay-embedding. Empirical insight into the characteristics of the noise effects will also be discussed.

\section{Numerical experiments on the noise effects}

\subsection{Two examples}

To investigate how stochasticity affects the partial observation, we perform numerical experiments. As shown in \cite{LinYT2021}, the data-driven approach enables us to obtain the Markov transition matrix and the memory kernel in \eref{eq_time_dependent_MZ_final} for deterministic cases, and the relationship between the memory kernel and the delay-embedding is described in the generalized Langevin equation. However, as discussed above,  it is not straightforward in stochastic cases to connect the memory kernel in \eref{eq_c_MZ_final_2} with the delay-embedding. Hence, we investigate the effects of the delay-embedding with the data-driven approach based on the EDMD algorithm in Sec.~2.3.

For deterministic systems, it is common that there are many variables, and we observe only a few of them. For the stochastic systems here, we assume that the noise term includes most of the unobserved effects. As discussed below, it is necessary to consider higher-order dictionary functions to evaluate the higher-order statistics in stochastic systems, which leads to high computational costs. Hence, as demonstrations, we examine the following two examples and the effects of partial observation.

The first one is the van der Pol system \cite{van_der_Pol1926} with additive noises:
\begin{eqnarray}
\rmd \underline{X_1} = \underline{X_2} \rmd t + \sigma_1 \underline{dW_1(t)},  \label{eq_vdP_1}\\
\rmd \underline{X_2} = \left( \mu \left( 1 - \underline{X_1^2}\right) \underline{X_2} - \underline{X_1} \right) \rmd t + \sigma_2 \underline{dW_2(t)}. \label{eq_vdP_2}
\end{eqnarray}
The second one is the Lorenz system \cite{Lorenz1963} with additive noises:
\begin{eqnarray}
\rmd \underline{X_1} = \nu \left( \underline{X_2} - \underline{X_1}\right) \rmd t + \sigma_1 \underline{dW_1(t)}, \label{eq_Lorenz_1}\\
\rmd \underline{X_2} = \left( \underline{X_1} \left( \rho - \underline{X_3}\right) - \underline{X_2} \right) \rmd t 
+ \sigma_2 \underline{dW_2(t)}, \label{eq_Lorenz_2}\\
\rmd \underline{X_3} = \left(\underline{X_1} \underline{X_2} - \beta \underline{X_3}\right) \rmd t + \sigma_3 \underline{dW_3(t)}. \label{eq_Lorenz_3}
\end{eqnarray}
In all the numerical experiments, we assume that the noise amplitude for each variable is common: $\sigma \equiv \sigma_1 = \sigma_2$ for the van der Pol system, and $\sigma \equiv \sigma_1 = \sigma_2 = \sigma_3$ for the Lorenz system. The system parameters are set to $\mu = 1$ for the van der Pol system, and $\nu = 10$, $\beta = 8/3$, and $\rho = 28$ or $13$ for the Lorenz system; we tried two values for the parameter $\rho$, as discussed below. Note that both systems are in periodic states above the Hopf bifurcations.

\begin{figure}[t]
\begin{center}
\includegraphics[width=120mm]{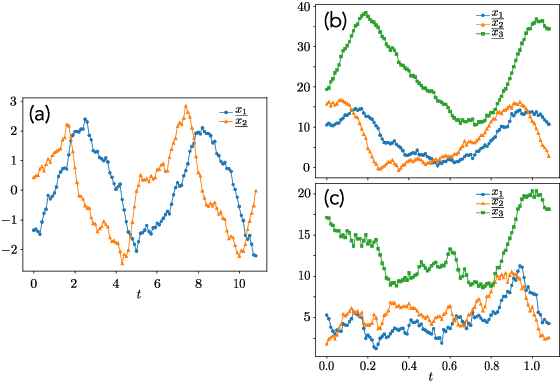}
\end{center}
\caption{Examples of sampled trajectories. (a) is for the van der Pol system with the noise amplitude $\sigma = 0.5$. (b) and (c) correspond to the  Lorenz cases with the noise amplitude $\sigma = 5$; we set $\rho = 28$ for (b) and $\rho = 13$ for (c). The markers indicate the observed data used in the EDMD algorithm. The intervals of the snapshot pairs are $\Delta t_{\mathrm{obs}} = 0.1$ for (a) and $\Delta t_{\mathrm{obs}} = 0.01$ for (b) and (c).}
\label{fig2}
\end{figure}

Figure~\ref{fig2} shows examples of sampled trajectories. The Euler-Maruyama approximation is employed \cite{Kloeden_book}; we set $\Delta t = 10^{-3}$ for the van der Pol system, and $\Delta t = 10^{-4}$ for the Lorenz system. Figure~\ref{fig2}(a) corresponds to the van der Pol system with the noise amplitude $\sigma = 0.5$; we see the noisy behavior along the limit cycle. Figures~\ref{fig2}(b) and \ref{fig2}(c) correspond to the Lorenz system with noise amplitude $\sigma = 5$. The difference between Figs.~\ref{fig2}(b) and \ref{fig2}(c) is in the system parameter $\rho$; $\rho = 28$ in Fig.~\ref{fig2}(b) and $\rho = 13$ in Fig.~\ref{fig2}(c). Although the Monte Carlo simulations are performed with the small time interval $\Delta t$, we construct the snapshot pairs from the generated data. Here, we set the time intervals as $\Delta t_{\mathrm{obs}} = 0.1$ for the van der Pol system and $\Delta t_{\mathrm{obs}} = 0.01$ for the Lorenz system. The markers in Fig.~\ref{fig2} are the observed data points.

\subsection{Evaluation of approximate values of exact solutions}

In deterministic systems, it is easy to obtain numerically exact solutions, which is necessary to discuss the accuracy of the partial observation. That is, the numerical time integration, such as the Runge-Kutta method, yields a good numerical approximation for the exact solutions. By contrast, the noise in the stochastic system prevents us from such numerical time integration. Although the Monte Carlo algorithm based on the Euler-Maruyama approximation is available for time integration, we should generate many sample trajectories to estimate the exact solutions for statistics. The problem here is the evaluations with \textit{various initial coordinates}; for each initial coordinate, we require many sample trajectories. Then, the Monte Carlo sampling is time-consuming. By contrast, as discussed in Sec.~2.4, we obtain the expectation values for various initial coordinates simply with the aid of the Koopman operator. That is, multiplying the corresponding Koopman matrix by the dictionary vector yields the expectation values. The numerical procedure is essentially the same as \cite{Ohkubo2019,Takahashi2023,Ohkubo2021}; we explain it briefly in the context of the present paper.

For example, consider three dimensional cases such as the Lorenz system. Then, we define the dictionary in Sec.~2.3 as follows:
\begin{eqnarray}
\bm{\psi} = [1, x_1, x_2, x_3, x_1^2, x_1 x_2, x_1 x_3, x_2^2, x_2 x_3, x_3^2, \dots]^{\top}.
\label{eq_dictionary_example}
\end{eqnarray}
Note that \eref{eq_c_time_evolution} leads to the expansion coefficients for the observable function $\phi$. Hence, when we can observe all variables, i.e., $\mathcal{O} = \{1,2,3\}$, we have the following expansion:
\begin{eqnarray}
\fl
x_1(\underline{\bm{x}}, t+\Delta t_{\mathrm{obs}}) \nonumber \\
\fl
= c_1^{x_1,\overline{\mathcal{O}}}(t+\Delta t_{\mathrm{obs}}) 1(\underline{\bm{x}})
+ c_1^{x_1,\mathcal{O}}(t+\Delta t_{\mathrm{obs}}) x_1(\underline{\bm{x}}) 
+ c_2^{x_1,\mathcal{O}}(t+\Delta t_{\mathrm{obs}}) x_2(\underline{\bm{x}}) \nonumber \\
\quad \fl
+ c_3^{x_1,\mathcal{O}}(t+\Delta t_{\mathrm{obs}}) x_3(\underline{\bm{x}}) 
 + \sum_{i \ge 2} c_i^{x_1, \overline{\mathcal{O}}}(t+\Delta t_{\mathrm{obs}}) \psi_i(\underline{\bm{x}}),
\label{eq_numerical_c_x_1}
\end{eqnarray}
where $\{c_d^{x_1,\mathcal{O}}(t+\Delta t_{\mathrm{obs}})\}$ and $\{c_i^{x_1, \overline{\mathcal{O}}}(t+\Delta t_{\mathrm{obs}})\}$ are obtained via the time integration of \eref{eq_c_time_evolution} with $\phi = x_1$. Next, we perform the time integration of \eref{eq_c_time_evolution} with $\phi = x_2$, $\phi = x_3$ and so on. Here, define $K^{\mathrm{full}}_{\mathrm{\Delta t_{\mathrm{obs}}}}$ as the Koopman matrix for the case where one can observe all the state variables. Then, the action of $K^{\mathrm{full}}_{\mathrm{\Delta t_{\mathrm{obs}}}}$ to $\bm{\psi}(\underline{\bm{x}},t)$ yields $\bm{\psi}(\underline{\bm{x}},t+\Delta t_{\mathrm{obs}})$; this indicates that the linear combination in \eref{eq_numerical_c_x_1} is directly related to the Koopman matrix $K^{\mathrm{full}}_{\Delta t_{\mathrm{obs}}}$. That is, the elements in the second row of the Koopman matrix $K^{\mathrm{full}}_{\mathrm{\Delta t_{\mathrm{obs}}}}$ correspond to the expansion coefficients $\{c_d^{x_1,\mathcal{O}}(t+\Delta t_{\mathrm{obs}})\}$ and $\{c_i^{x_1, \overline{\mathcal{O}}}(t+\Delta t_{\mathrm{obs}})\}$. Similarly, the third and fourth rows of $K^{\mathrm{full}}_{\Delta t_{\mathrm{obs}}}$ correspond to the coefficients of $x_2$ and $x_3$, respectively. 

Repeating this procedure, it is possible to evaluate the Koopman matrix $K^{\mathrm{full}}_{\Delta t_{\mathrm{obs}}}$ numerically. In the numerical experiments, we employ the Crank-Nicolson method to solve \eref{eq_c_time_evolution}; the discrete time intervals for numerical time integration are set to $\Delta t = 10^{-3}$ for the van der Pol system and $\Delta t = 10^{-4}$ for the Lorenz system, respectively. As denoted above, we set the observation time interval for the snapshot pairs as $\Delta t_{\mathrm{obs}} = 0.1$ for the van der Pol system and $\Delta t_{\mathrm{obs}} = 0.01$ for the Lorenz system. The maximum degrees in the dictionary are $8$ for the van der Pol system and $6$ for the Lorenz system; we confirmed that these dictionary function settings provide sufficient accuracy in the absence of noise. Finally, we obtain the Koopman matrix $K^{\mathrm{full}}_{\Delta t_{\mathrm{obs}}}$.

As discussed in Sec.~2.4, the Koopman matrix $K^{\mathrm{full}}_{\Delta t_{\mathrm{obs}}}$ yields the expectation values after the time-evolution with $\Delta t_{\mathrm{obs}}$; see \eref{eq_action_Koopman_basic}. Here, we consider the coordinate $\underline{\bm{x}}$. Then, the action of $K^{\mathrm{full}}_{\Delta t_{\mathrm{obs}}}$ on the dictionary $\bm{\psi}(\underline{\bm{x}})$ yields the following vector:
\begin{eqnarray}
\left[
\begin{array}{c}
\mathbb{E}\left[
\psi_1(\underline{\bm{x}(t+\Delta t_{\mathrm{obs}})})
\big| \underline{\bm{x}(t)} = \underline{\bm{x}}
\right] \\
\mathbb{E}\left[
\psi_2(\underline{\bm{x}(t+\Delta t_{\mathrm{obs}})})
\big| \underline{\bm{x}(t)} = \underline{\bm{x}}
\right] \\
\vdots 
\end{array}
\right]
= K^{\mathrm{full}}_{\Delta t_{\mathrm{obs}}} \bm{\psi}(\underline{\bm{x}}).
\end{eqnarray}
When the dictionary function in \eref{eq_dictionary_example} is employed, the following expectation values are evaluated by the corresponding vector elements of $K^{\mathrm{full}}_{\Delta t_{\mathrm{obs}}} \bm{\psi}(\underline{\bm{x}})$:
\begin{eqnarray}
\mathbb{E}[\underline{X_1}] \leftrightarrow \left[K^{\mathrm{full}}_{\Delta t_{\mathrm{obs}}} \bm{\psi}(\underline{\bm{x}})\right]_{2}, \nonumber \\
\mathbb{E}[\underline{X_2}] \leftrightarrow \left[K^{\mathrm{full}}_{\Delta t_{\mathrm{obs}}} \bm{\psi}(\underline{\bm{x}})\right]_{3}, \nonumber \\
\mathbb{E}[\underline{X_3}] \leftrightarrow \left[K^{\mathrm{full}}_{\Delta t_{\mathrm{obs}}} \bm{\psi}(\underline{\bm{x}})\right]_{4}. \nonumber
\end{eqnarray}
The higher order statistics are also evaluated as follows:
\begin{eqnarray}
\mathbb{E}[\underline{X_1^2}] \leftrightarrow \left[K^{\mathrm{full}}_{\Delta t_{\mathrm{obs}}} \bm{\psi}(\underline{\bm{x}})\right]_{5}, \nonumber \\
\mathbb{E}[\underline{X_1 X_2}] \leftrightarrow \left[K^{\mathrm{full}}_{\Delta t_{\mathrm{obs}}} \bm{\psi}(\underline{\bm{x}})\right]_{6}, \nonumber 
\end{eqnarray}
and so on.

As described above, once we obtain the Koopman matrix $K^{\mathrm{full}}_{\Delta t_{\mathrm{obs}}}$, it is possible to calculate the time-evolved expectation values for various initial coordinates. That is, only the multiplication of $K^{\mathrm{full}}_{\Delta t_{\mathrm{obs}}}$ by an initial vector yields the expectation values. In the following numerical experiments, we use the calculated values by $K^{\mathrm{full}}_{\Delta t_{\mathrm{obs}}}$ as the exact solutions. Then, we evaluate absolute errors between values calculated by partial observation and the exact ones.

\subsection{Settings for the EDMD algorithm for partial observation}

To investigate the effects of the partial observation, we employ the EDMD algorithm with the delay-embedding. The EDMD algorithm is data-driven, and we generate the dataset as follows: 
\begin{enumerate}
\item Generate randomly initial coordinates. (For the van der Pol system, a uniform density with $[-1,+1]^2$ was used. For the Lorenz system, we used a uniform density with $[-10,+10]^3$.)
\item Using the Euler-Maruyama approximation, the time integration is simulated via the Monte Carlo method. The time intervals for time-evolution and observation are the same as those in Sec.~5.1.
\item After $100 \times \Delta t_{\mathrm{obs}}$ relaxation time, we take $101 + M_{\mathrm{max}}$ data points, where $M_{\mathrm{max}}$ is the maximum number of the delay-embedding. 
\item The above procedures are repeated $100$ times. 
\end{enumerate}
We prepare $10,000$ snapshot pairs and construct the Koopman matrix from the dataset. Although data points for all coordinates are generated, we consider the partial observation with only one variable.
 
We should note the dictionary in the EDMD algorithm. In deterministic systems, it is common to use a naive delay-embedding of the past coordinate, such as $x_d(t-\Delta t_{\mathrm{obs}})$. Although this setting is enough to evaluate only the coordinate values, we should employ higher-order basis functions to evaluate higher-order statistics. That is, $\mathbb{E}[\underline{X_d^2}] \neq (\mathbb{E}[\underline{X_d}])^2$ in the stochastic systems, and the evaluation of $\mathbb{E}[\underline{X_d^2}]$ requires the monomial functions whose maximum degree is at least two.

To consider the higher-order basis functions with the delay-embedding, we rewrite the variables $\underline{\bm{x}}$ in Sec.~2 with $\underline{\bm{z}}$. When we observe the $d$-th variable, we set
\begin{eqnarray}
\left[
\begin{array}{c}
\underline{z_1(t)} \\
\underline{z_2(t)} \\
\vdots \\
\underline{z_{M+1}(t)}
\end{array}
\right]
=
\left[
\begin{array}{c}
\underline{x_d(t)} \\
\underline{x_d(t-\Delta t_{\mathrm{obs}})}\\
\vdots\\
\underline{x_d(t-M\Delta t_{\mathrm{obs}})}
\end{array}
\right],
\label{eq_variable_z}
\end{eqnarray}
where $M$ is the number of the delay-embedding. Then, we introduce higher-order monomial basis functions for $\{z_i(t)\}$ to evaluate higher-order statistics. Since we will focus on the mean values and the variances, the maximum degree of the dictionary functions for $\{z_i(t)\}$ is set to two, which leads to the following dictionary $\bm{\psi}^{\mathrm{part}}$:
\begin{eqnarray}
\fl
[1, z_1, \dots, z_{M+1}, z_1^2, z_1 z_2, z_1 z_3, \dots, z_{M}z_{M+1}, z_{M+1}^2] \nonumber \\
\fl
=[1, x_d(t), \dots, x_d(t-M\Delta t_{\mathrm{obs}}),  x_d(t)^2, x_d(t) x_d(t-\Delta t_{\mathrm{obs}}), x_d(t) x_d(t-2\Delta t_{\mathrm{obs}}), \nonumber \\
\fl \quad\,\,
\dots, x_d(t-(M-1)\Delta t_{\mathrm{obs}}) x_d(t-M\Delta t_{\mathrm{obs}}), x_d(t-M\Delta t_{\mathrm{obs}})^2],
\label{eq_dictionary_z_2nd}
\end{eqnarray}
where $x_d(t-m\Delta t_{\mathrm{obs}})$ is a function which yields the coordinate value in the $m \Delta t_{\mathrm{obs}}$ past:
\begin{eqnarray}
x_d(t-m\Delta t_{\mathrm{obs}})(\underline{\bm{z}(t)}) = \underline{x_d(t-m\Delta t_{\mathrm{obs}})}.
\end{eqnarray}

By using the delay-embedded vector $\underline{\bm{z}}$ and the dictionary $\bm{\psi}^{\mathrm{part}}$, it is possible to construct the Koopman matrix $K^{\mathrm{part}}_{\Delta t_{\mathrm{obs}}}$ in a similar way to that of Sec.~5.2. Note that the partial observation leads to the Koopman matrix $K^{\mathrm{part}}_{\Delta t_{\mathrm{obs}}}$, and approximate expectation values after the time-evolution with $\Delta t_{\mathrm{obs}}$ are evaluated by $K^{\mathrm{part}}_{\Delta t_{\mathrm{obs}}} \bm{\psi}^{\mathrm{part}}(\underline{\bm{z}})$. Then, the following procedure is employed to investigate the effects of the partial observation.

First, we generate $N_{\mathrm{test}}$ initial coordinates randomly; uniform distributions on $[-1,+1]^2$ for the van der Pol system and $[-10,+10]^3$ for the Lorenz system are employed. The $n$-th generated coordinate is denoted as $\underline{\bm{x}^{(n)}}$. Then, the exact expectation values after the time-evolution with $\Delta t_{\mathrm{obs}}$ are given by $K^{\mathrm{full}}_{\Delta t_{\mathrm{obs}}} \bm{\psi}(\underline{\bm{x}^{(n)}})$. Next, we only observe a coordinate $x_d^{(n)}$ from $\bm{x}^{(n)}$ and construct the delay-embedded coordinate $\underline{\bm{z}^{(n)}}$. The corresponding expectation values are evaluated by $K^{\mathrm{part}}_{\Delta t_{\mathrm{obs}}} \bm{\psi}^{\mathrm{part}}(\underline{\bm{z}^{(n)}})$. Assume the $d$-th element of $K^{\mathrm{full}}_{\Delta t_{\mathrm{obs}}} \bm{\psi}(\underline{\bm{x}^{(n)}})$ and $d'$-th element of  $K^{\mathrm{part}}_{\Delta t_{\mathrm{obs}}} \bm{\psi}^{\mathrm{part}}(\underline{\bm{z}^{(n)}})$ correspond to the target statistics, respectively. Then, the error is defined as 
\begin{eqnarray}
\mathrm{error} = 
\frac{1}{N_{\mathrm{test}}}
\sum_{n=1}^{N_{\mathrm{test}}}
\left|
\left[K^{\mathrm{full}}_{\Delta t_{\mathrm{obs}}} \bm{\psi}\left( \underline{\bm{x}^{(n)}} \right) \right]_d
- 
\left[K^{\mathrm{part}}_{\Delta t_{\mathrm{obs}}} \bm{\psi}\left( \underline{\bm{z}^{(n)}} \right) \right]_{d'}
\right|.
\end{eqnarray}
In the following experiments, we set $N_{\mathrm{test}} = 1000$.

\subsection{Comparison with higher-order dictionary functions and delay-embedding}

\begin{figure}[t]
\begin{center}
\includegraphics[width=120mm]{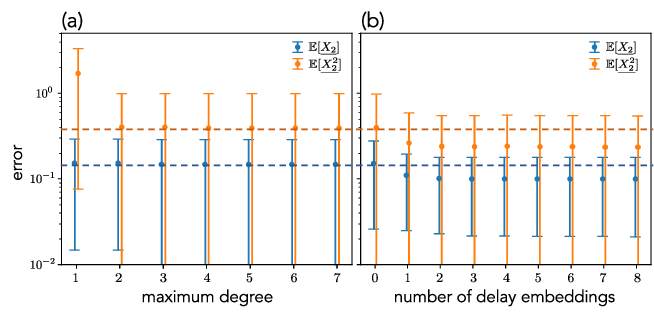}
\end{center}
\caption{Errors of the estimated statistics by the Koopman matrix with the partial observation in the van der Pol system with the noise amplitude $\sigma = 0.5$. (a) Dependence on the choice of dictionary functions without delay-embedding. (b) Dependence of the number of delay-embeddings. For the sake of visibility, the results for $\mathbb{E}[\underline{X_2}]$ and $\mathbb{E}[\underline{X_2^2}]$ are drawn slightly shifted. The horizontal dashed lines are eye guides.}
\label{fig3}
\end{figure}

As discussed in Sec.~4.2, the delay-embedding would improve the accuracy of statistics in the Koopman operator approach. Here, we present the numerical results for the van der Pol system with the noise amplitude $\sigma = 0.5$.

Figure~\ref{fig3} shows the errors in the partial observation setting. Here, only the second variable, $\underline{x_2}$, is measurable. The filled circles and error bars represent the mean values and the standard deviations of the errors, respectively. In Fig.~\ref{fig3}(a), we do not use the delay-embeddings, and the maximum degree of the basis functions is changed. We see that the sudden change occurs when the maximum degree is varied from 1 to 2; this fact means that the monomial functions with degree 2 are necessary to evaluate the second-order statistics.

Figure~\ref{fig3}(b) shows the relationship between the number of delay-embedding and error. Note that the dictionary introduced in Section 5.3 appropriately estimates $\mathbb{E}[\underline{X_2^2}]$. The numerical results indicate that increasing the number of delay-embedding improves accuracy. It also confirms the existence of a saturation point.

Compared to Figs.~\ref{fig3}(a) and \ref{fig3}(b), we confirmed that delay-embedding is more effective than using higher-order dictionary functions; the delay embedding yields a slightly better accuracy than the higher-order dictionaries. This finding holds for all other cases involving different noise amplitudes and Lorenz systems. Therefore, the discussion in Sec.~4.2 would be valid: Delay-embedding is also effective in the partially observed case within stochastic systems.

\subsection{Effects of higher order dictionary functions}

\begin{figure}[t]
\begin{center}
\includegraphics[width=120mm]{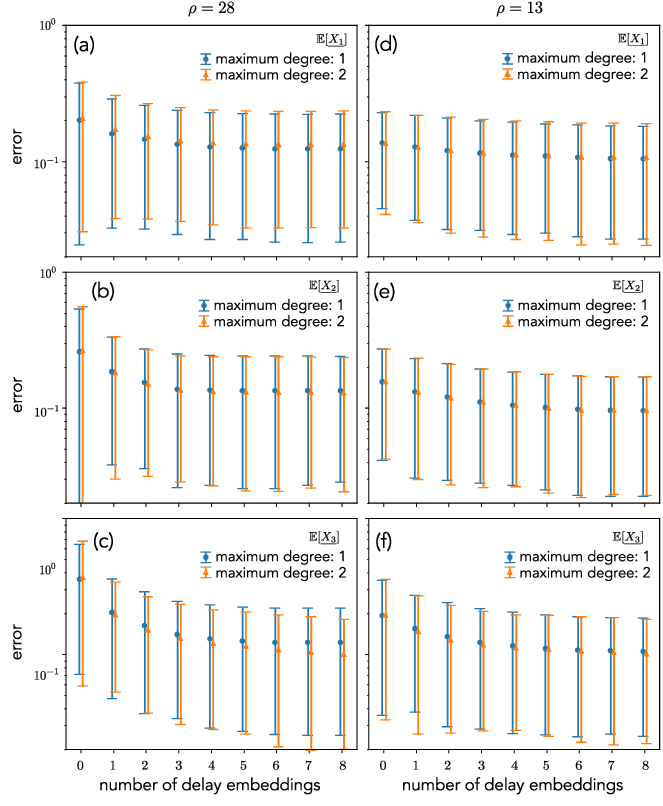}
\end{center}
\caption{Errors of the first-order statistics in the partial observation settings for the Lorenz system with the noise amplitude $\sigma = 3$. In (a) and (d), only the first variable, $\underline{x_1}$, is observable; $\underline{x_2}$ for (b) and (e), and $\underline{x_3}$ for (c) and (f). (a), (b), and (c) corresponds to the cases with the system parameter $\rho = 28$; (d), (e), and (f) for $\rho = 13$.}
\label{fig4}
\end{figure}

As explained in Sec.~4.2, the use of the higher-order basis functions is necessary for the evaluation of higher-order statistics. We here investigate its effects on the first-order statistics, i.e., the mean values. In the numerical experiments, we show the results for the Lorenz system. Note that we checked that the same behavior occurs in the van der Pol system. The noise amplitude for the Lorenz system is set to $\sigma=3$.

Figure~\ref{fig4} shows the errors of the first-order statistics in the partial observation settings. We compare two dictionaries for the delay-embeddings; one is the second-order monomial basis functions for $\{z_i\}$ in \eref{eq_dictionary_z_2nd}. The other one is the first-order monomial basis functions. Only in the case of $\mathbb{E}[\underline{X_3}]$, as shown in Fig.~\ref{fig4}(c), there are slight variations in error for different dictionaries. The reason for this would be as follows. In the Lorenz system, $\underline{X_3}$ differs from $\underline{X_1}$ and $\underline{X_2}$; \eref{eq_Lorenz_1} does not include a term related to $\underline{X_3}$, while each of the three equations includes $\underline{X_1}$ and $\underline{X_2}$. Hence, the subspace spanned by the delay-embedding with $\underline{X_3}$  would be smaller than those with $\underline{X_1}$ and $\underline{X_2}$. Then, the difference in the dictionary affects the partial observation of $\underline{X_3}$. As $\rho$ becomes smaller, the effects of noise through $\underline{X_2}$  could be smaller. Hence, the effects on partial observations also become smaller; compared to Fig.~\ref{fig4}(c), the difference in error between different dictionaries in Fig.~\ref{fig4}(f) is smaller. From these results, we conclude that the first-order basis functions with delay-embeddings contain enough information for the time-evolution in the function space. The discussion in Sec.~4.2 justifies the results; the simple delayed function contains the information of the time-evolution; see  \eref{eq_delay_explanation_2}.

Of course, note that the evaluation of the second-order statistics, such as $\mathbb{E}[\underline{X_d^2}]$, requires the second-order monomial basis functions.

\subsection{Power-law behavior for the change of noise amplitude}

In general, external parameters such as temperature control the amplitude of the additive noise. Here, we investigate the effects of the noise amplitude. As a consequence, we found a power-law dependency of the errors on the noise amplitude.

\begin{figure}[t]
\begin{center}
\includegraphics[width=100mm]{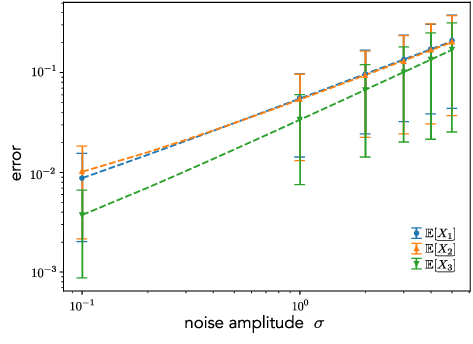}
\end{center}
\caption{The dependency of errors on the noise amplitude in the Lorenz system with $\rho = 28$. The dashed lines correspond to the fitting results with the function form in \eref{eq_fitting_function}.}
\label{fig5}
\end{figure}

Figure~\ref{fig5} depicts the dependency of the errors on the noise amplitude $\sigma$ in the Lorenz system. Here, the plotted markers correspond to the errors for the delay-embeddings with $M = 8$; for all cases, the errors seem to have a certain saturation point in Fig.~\ref{fig4}. Hence, we judged that the $M=8$ cases yield an accurate estimation within the finite dictionary functions and used them in Fig.~\ref{fig5}.

Note that we here employ a log-log plot in Fig.~\ref{fig5}. From the plotted results, we assume the function form of the errors with respect to the noise amplitude as follows:
\begin{eqnarray}
f(\sigma) = \alpha_2 \sigma^{\alpha_1} + \alpha_3.
\label{eq_fitting_function}
\end{eqnarray}
Even in the noiseless cases, i.e., deterministic systems, there is a difference between the exact values and the estimated ones because of the partial observation. Hence, we introduced $\alpha_3$ in \eref{eq_fitting_function}. The fitted results are also shown in Fig.~\ref{fig5}; the fitting seems to be good.

\begin{table}
\caption{The power-law exponent $\alpha_1$ in \eref{eq_fitting_function} for the Lorenz system.}
\center
\begin{tabular}{cccc}
\hline\hline
       & $\alpha_1$ for $\mathbb{E}[\underline{X_1}]$ & $\alpha_1$ for $\mathbb{E}[\underline{X_2}]$  & $\alpha_1$ for $\mathbb{E}[\underline{X_3}]$  \\
\hline
$\rho = 28$ & 0.808 $\pm$ 0.012 & 0.849 $\pm$ 0.022 & 1.007 $\pm$ 0.013\\
$\rho = 13$ & 0.773 $\pm$ 0.036 & 0.661 $\pm$ 0.024 & 0.820 $\pm$ 0.022 \\
\hline\hline
\end{tabular}
\label{table_exponents_1}
\end{table}

In Table~\ref{table_exponents_1}, we summarize the power-law exponent $\alpha_1$ in \eref{eq_fitting_function} for the Lorenz system. Here, we repeated the same procedure for Fig.~\ref{fig5} five times and evaluated the means and standard deviations for the exponent $\alpha_1$. Although it is difficult to capture the meaning of the exponent $\alpha_1$, the exponent would reflect the effect of the unobserved part of the partial observation. The reason is as follows:
\begin{itemize}
\item Although the EDMD algorithm can adequately deal with noise effects, partial observation causes an increase in error due to noise effects. The error remains unaffected only when the noise effects are orthogonal to the space generated by the observable function.
\item As discussed in Sec.~5.5, $\underline{X_3}$ is the most significantly affected by partial observations. Then, the exponent $\alpha_1$ for $\mathbb{E}[\underline{X_3}]$ is the largest in Table~\ref{table_exponents_1}. Note that the error values for $\mathbb{E}[\underline{X_3}]$ are the smallest in Fig.~\ref{fig5}; the increase in error due to the partial observation is not directly related to the actual error.
\item All the exponents $\alpha_1$ for $\rho=13$ are smaller than those for $\rho=28$. The reason is that the term $\rho \underline{X_1}$ in \eref{eq_Lorenz_2} becomes smaller, and the effects of the partial observation also diminish. In particular, the effects of noise in $\underline{X_1}$ are reduced directly via the term $\rho \underline{X_1}$, and $\underline{X_2}$ is most significantly affected.
\end{itemize}

From the above discussion, we expect that the power-law form and its exponent reflect some parts of the effects of the partial observation. If the ignored term has a strong nonlinearity, the impact of the partial observation should be considerable. Then, we perform additional numerical experiments for the following modified van der Pol systems, in which we replace \eref{eq_vdP_2} with
\begin{eqnarray}
\rmd \underline{X_2} = \left( \mu \left( 1 - \underline{X_1^2}\right) \underline{X_2} - h(\underline{X_1}) \right) \rmd t + \sigma_2 \underline{dW_2(t)},
\end{eqnarray}
where $h(\underline{X_1})$ is a nonlinear function. In the original van der Pol system, $h(\underline{X_1}) = \underline{X_1}$. This is an odd function, and then, we consider two cases with $h(\underline{X_1}) = \underline{X_1^3}$ and $h(\underline{X_1}) = \underline{X_1^5}$.

\begin{table}
\caption{The power-law exponent $\alpha_1$ in \eref{eq_fitting_function} for the modified van der Pol system.}
\center
\begin{tabular}{ccc}
\hline\hline
       & $\alpha_1$ for $\mathbb{E}[\underline{X_1}]$ & $\alpha_1$ for $\mathbb{E}[\underline{X_2}]$  \\
\hline
$h(\underline{X_1}) = \underline{X_1}$   & 0.489 $\pm$ 0.027 & 0.596 $\pm$ 0.017 \\
$h(\underline{X_1}) = \underline{X_1^3}$ & 0.567 $\pm$ 0.030 & 0.652 $\pm$ 0.029 \\
$h(\underline{X_1}) = \underline{X_1^5}$ & 0.794 $\pm$ 0.012 & 1.197 $\pm$ 0.102 \\
\hline\hline
\end{tabular}
\label{table_exponents_2}
\end{table}

Table~\ref{table_exponents_2} shows the numerical results. As expected from the above discussion, the stronger nonlinearity yields a larger exponent.

\section{Discussions}

In this work, we focused on the partial observations in stochastic systems. The Koopman operator approach is also available, and it is crucial to note the difference between state variables and observable functions. In other words, the Koopman operator approach gives an expected value, not the value of the state variable. The deterministic case is special, and the expected value coincides with the value of the state variable, i.e., $\mathbb{E}[\underline{X_d}] = \underline{x_d}$. Although the stochasticity prevents us from using this simple connection, the delay-embedding is effective even in stochastic systems. Note that there is a difference with the deterministic case, i.e., higher-order dictionary functions for the embedded states are necessary to evaluate higher-order statistics. In addition, the numerical experiments clarified that the noise dependency for error exhibits a power-law behavior, and the exponent could reflect the characteristics of the partial observations.

From a slightly different perspective, Baldovin \textit{et al.} discussed the effects of partial observation in stochastic systems \cite{Baldovin2020}. In \cite{Baldovin2020}, the authors investigated the relationship between correlation and causation; they conclude that the delay-embedding strategy based on Takens theorem does not work well in nonlinear stochastic systems. The present work could provide reasons why the embedding approach does not work in the correlation and causation discussion in \cite{Baldovin2020}. At least, we can say that the deterministic case is a special case from the perspective of the Mori-Zwanzig formulation and the Koopman operator approach; the time-evolution equation for the observable coincides with that for the original coordinate variable. This fact could explain why Takens theorem works well in deterministic systems but not in stochastic systems. While further studies are necessary, the discussion based on the Mori-Zwanzig formulation and the Koopman operator approach would contribute to related theoretical discussions in partial observations.

There are some remaining tasks. First, it is not clear why the dependency of error on the noise amplitude exhibits the power-law form in \eref{eq_fitting_function}. Second, one should investigate the theoretical connection between the exponent $\alpha_1$ in the power-law form and the information loss in partial observations. While we performed several other numerical experiments, we have not yet found a quantitative correspondence.

As mentioned above, there has been little discussion of partial observations in stochastic systems. Modeling based on stochastic differential equations is popular. Hence, further studies on partial observations are desirable from a practical standpoint. We believe that this study will serve as a starting point for future research.

\ack
This work was supported by JST FOREST Program (Grant Number JPMJFR216K, Japan).


\end{document}